\documentclass[journal]{IEEEtran}
\usepackage{cite}
\usepackage{amsmath,amssymb,amsfonts}
\usepackage{graphicx}
\usepackage{xcolor}
\usepackage[ruled,vlined]{algorithm2e}
\usepackage{multirow}
\usepackage{booktabs}
\usepackage{siunitx}
\usepackage[numbers,compress]{natbib}
\usepackage{bm,bbm}
\usepackage[hidelinks]{hyperref}
\ifCLASSINFOpdf

\else

\fi

\begin{document}
%
\title{SAM 3D for 3D Object Reconstruction from Remote Sensing Images}
%
%
%

\author{Junsheng~Yao, Lichao~Mou, and~Qingyu~Li*
\thanks{J. Yao and Q. Li are with The Chinese University of Hong Kong, Shenzhen, Longgang, Shenzhen, Guangdong, China (e-mail: junshengyao@link.cuhk.edu.cn; qingyu.li@cuhk.edu.cn).}
\thanks{L. Mou is with MedAI Technology (e-mail: lichao.mou@gmail.com).}
\thanks{Corresponding author: Qingyu Li.}}

\maketitle

\begin{abstract}
Monocular 3D building reconstruction from remote sensing imagery is essential for scalable urban modeling, yet existing methods often require task-specific architectures and intensive supervision. This paper presents the first systematic evaluation of SAM 3D, a general-purpose image-to-3D foundation model, for monocular remote-sensing building reconstruction. We benchmark SAM 3D against TRELLIS on the NYC urban dataset samples, employing Fr\'echet Inception Distance (FID) and CLIP-based Maximum Mean Discrepancy (CMMD) as evaluation metrics. Experimental results demonstrate that SAM 3D produces more coherent roof geometry and sharper boundaries compared to TRELLIS. We further extend SAM 3D to urban scene reconstruction through a \emph{segment--reconstruct--compose} pipeline, demonstrating its potential for urban scene modeling. We also analyze practical limitations and discuss future research directions. These findings provide practical guidance for deploying foundation models in urban 3D reconstruction and motivate future integration of scene-level structural priors.
\end{abstract}


%
\IEEEpeerreviewmaketitle

\section{Introduction}
%
%
%
%
\IEEEPARstart{A}{ccurate} 3D building models are a cornerstone of city-scale 3D modeling and digital-twin systems, and are critical for downstream applications such as urban analytics, simulation, and planning \cite{mazzetto2024review}. In practice, such 3D models can be obtained from different data sources, including multi-view imagery, LiDAR point clouds, and monocular remote-sensing images \cite{luo2024large}. While multi-view and LiDAR-based pipelines often provide strong geometric cues, they can be costly to acquire, spatially incomplete, or difficult to scale consistently across large areas \cite{luo2024large}. Monocular remote-sensing reconstruction, which infers 3D building geometry from a single image, is therefore attractive for scalable urban modeling; however, it remains intrinsically challenging due to strong top-down viewpoints, repetitive rooftop patterns, weak façade evidence, and complex shadows or occlusions \cite{li2024sat2scene}. Existing monocular building reconstruction methods typically rely on task-specific architectures and intensive supervision (e.g., multi-task geometric parsing or multi-level annotations) \cite{shi2025instance}, which may incur high annotation cost and suffer reduced robustness when transferred across cities or imaging conditions \cite{li20213d, li20243d}.

Recent progress in foundation models suggests a complementary direction. General-purpose segmentation models such as Segment Anything (SAM) have demonstrated strong instance delineation across diverse visual domains \cite{kirillov2023segment}. In parallel, generative image-to-3D foundation models have shown encouraging open-world reconstruction capabilities from a single image. TRELLIS introduces a structured latent representation that enables scalable 3D generation and flexible decoding to multiple output formats \cite{xiang2025structured}. More recently, SAM 3D emphasizes visually grounded single-image 3D reconstruction with improved robustness in unconstrained natural images \cite{chen2025sam}. However, whether such general-purpose models can be reliably applied to monocular remote sensing imagery remains underexplored, given the substantial domain gap in both appearance statistics and geometric priors (e.g., roof-dominant cues, weak façade signals, and highly repetitive urban layouts). To our knowledge, a systematic evaluation of SAM 3D for monocular remote-sensing building reconstruction has not been reported.

The goal of this research is to investigate the effectiveness of SAM 3D in the task of 3D building reconstruction from monocular remote sensing images. Our work presents three main contributions:

(i) We provide the first systematic evaluation of SAM 3D for monocular remote-sensing 3D building reconstruction, benchmarking it against TRELLIS using Fr\'echet Inception Distance (FID)~\cite{heusel2017gans} and the CLIP-based CMMD metric~\cite{jayasumana2024rethinking}, discovering SAM 3D's superiority in remote-sensing single-building modeling.

(ii) We extend SAM 3D to urban scene reconstruction via a \emph{segment--reconstruct--compose} pipeline, providing an initial validation of general-purpose image-to-3D foundation models for multi-building reconstruction.

(iii) We discuss practical applicability, limitations, and future directions, motivating the integration of instance-level foundation models with scene-level structural priors for robust urban reconstruction.

\section{Methodology}
\begin{figure*}[!t]
  \centering
  \includegraphics[width=1.0\textwidth]{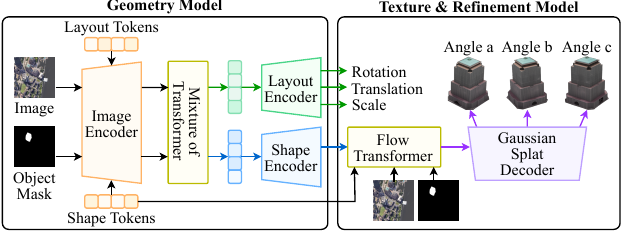}
  \caption{SAM 3D architecture for evaluation-driven 3D building reconstruction from monocular remote sensing imagery.}
  \label{fig_1}
\end{figure*}

Given a monocular remote sensing RGB image $I \in \mathbb{R}^{H \times W \times 3}$, we obtain a target-building binary instance mask $M \in\{0,1\}^{H \times W}$ using an interactive tool (EISeg). We then reconstruct a renderable 3D building asset conditioned on $(I, M)$. To enable scene-level reconstruction, we output an object-centric yet composable representation consisting of geometry and a camera-frame layout, parameterized by rotation $R$, translation $t$, and scale $s$.

As illustrated in Fig.~\ref{fig_1}, we feed $(I, M)$ into SAM 3D, which operates in two stages: (i) a Geometry Model that jointly predicts a coarse 3D shape and an object layout, and (ii) a Texture \& Refinement Model that enriches high-frequency geometry and appearance on active voxels. The final output is exported for rendering-based evaluation and visualization.

To capture both roof-level details and global context cues, SAM 3D encodes an object-centric crop and the full image view, and uses the resulting conditioning tokens throughout generation. In the geometry stage, the model predicts a coarse voxel latent $O \in \mathbb{R}^{64^3}$ together with $(R, t, s)$. A Mixture-of-Transformers (MoT) couples shape and layout tokens via a structured self-attention mask, improving shape–layout consistency. In the refinement stage, a sparse latent flow Transformer refines geometry and synthesizes appearance, and a Gaussian-splat decoder produces a continuous renderable representation. 

For urban scene reconstruction, we apply a \emph{segment--reconstruct--compose} pipeline: each building instance is reconstructed independently, and the resulting Gaussian Splatting representations are merged into a unified coordinate frame using the predicted layout parameters. This strategy leverages SAM 3D's composable output representation, enabling the extension of single-object reconstruction to urban scene modeling.

\section{Experiments}
\subsection{Dataset}
All experiments are conducted on the NYC urban remote sensing dataset introduced by CityDreamer~\cite{xie2024citydreamer}, which has been widely adopted as a benchmark in subsequent works such as Skyfall-GS~\cite{lee2025skyfall}. The dataset covers Manhattan and Brooklyn, consisting of high-resolution remote sensing images paired with a consistent city-scale 3D scene representation rendered using Google Earth Studio. 

Images are rendered from high elevation viewpoints (approximately 80°) to approximate satellite imaging conditions, reflecting typical challenges in remote sensing scenarios such as limited parallax and severe facade occlusions. We use these remote sensing images as monocular input, and the corresponding Google Earth Studio renderings from alternative viewpoints serve as approximate ground truth for evaluating reconstruction quality at the image distribution level.

From the dataset, we select representative urban regions and further extract a set of standalone building instances for evaluation. Each building instance is associated with a single remote sensing image as input.

\subsection{Experimental setup}
For building-level reconstruction, SAM 3D and TRELLIS are applied to selected building instances. Each building is reconstructed from a single remote sensing image, producing a 3D Gaussian Splatting representation exported under identical input conditions. For quantitative evaluation, reconstructed representations are rendered offline from multiple unobstructed viewpoints. Rendered images are aligned with corresponding reference views for metric computation, and the final score for each building is obtained by averaging across all viewpoints.

For urban scene reconstruction, we apply the \emph{segment--reconstruct--compose} pipeline to reconstruct and merge all buildings within a single scene image. The reconstruction quality is assessed through visual comparison against the ground truth.

\subsection{Evaluation Metrics}
We employ two distribution-based perceptual metrics to quantitatively evaluate 3D building reconstruction under monocular remote sensing imagery: Fréchet Inception Distance (FID) and CLIP-based Maximum Mean Discrepancy (CMMD). Both metrics are computed by comparing multi-view renderings of reconstructed models with corresponding reference images in learned feature spaces.

FID measures the Fréchet distance between feature distributions of generated and reference images in the Inception-v3 feature space. Let $\boldsymbol{\mu}_r, \boldsymbol{\Sigma}_r$ and $\boldsymbol{\mu}_g, \boldsymbol{\Sigma}_g$ denote the mean vectors and covariance matrices of reference and generated image features, respectively. FID is defined as
\begin{equation}
\mathrm{FID}=\left\|\boldsymbol{\mu}_r-\boldsymbol{\mu}_g\right\|_2^2+\operatorname{Tr}\left(\boldsymbol{\Sigma}_r+\boldsymbol{\Sigma}_g-2\left(\boldsymbol{\Sigma}_r \boldsymbol{\Sigma}_g\right)^{1 / 2}\right)
\end{equation}
Lower FID values indicate closer alignment between the generated and reference image distributions.

To complement FID with a semantic-aware criterion, we further adopt Maximum Mean Discrepancy computed in the CLIP feature space (CMMD). Image features are extracted and $\ell_2$-normalized using a pretrained CLIP ViT-L/14 model. The squared MMD with a radial basis function (RBF) kernel is given by
\begin{equation}
\operatorname{MMD}^2=\mathbb{E}_{x, x^{\prime}}\left[k\left(x, x^{\prime}\right)\right]+\mathbb{E}_{y, y^{\prime}}\left[k\left(y, y^{\prime}\right)\right]-2 \mathbb{E}_{x, y}[k(x, y)]
\end{equation}

where $k(x, y)=\exp \left(-\gamma\|x-y\|_2^2\right)$, and the bandwidth parameter $\gamma$ is adaptively estimated using the median heuristic. Lower CMMD values correspond to higher semantic consistency between reconstructed and reference images.

Both FID and CMMD are computed independently for each rendered viewpoint, and the final metric for a building is obtained by averaging across all viewpoints to ensure robustness against viewpoint variations.

\section{Results}
\subsection{Quantitative Comparison on Individual Buildings}
\begin{figure*}[!ht]
  \centering
  \includegraphics[width=1.0\textwidth]{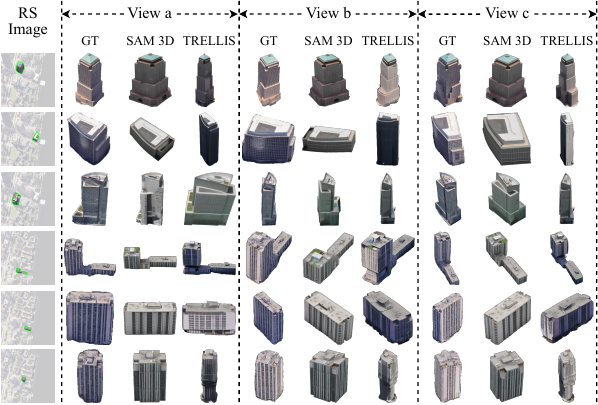}
  \caption{Visual comparison of reconstructed buildings. Each row corresponds to one building. Columns 1/2 show the input remote-sensing image and the corresponding GT view; columns 3--4 show SAM 3D and TRELLIS results. Columns 5--10 repeat the GT/SAM 3D/TRELLIS triplets for two additional viewpoints.}
  \label{fig_2}
\end{figure*}

To comprehensively evaluate reconstruction quality, we render and assess each building from three distinct viewpoints. Fig.~\ref{fig_2} shows qualitative comparisons across these viewpoints. SAM 3D generally reconstructs more coherent building shapes, with sharper boundaries and noticeably improved roof geometry. In particular, rooftop structures produced by SAM 3D better match the reference views, while TRELLIS tends to produce oversmoothed roofs or local shape distortions.

Table~\ref{tab:quantitative} reports the averaged FID and CMMD across the three viewpoints for each building. SAM 3D achieves lower average FID and CMMD than TRELLIS, with an overall reduction of about 8\% in FID and 18\% in CMMD. SAM 3D performs particularly better on buildings with complex rooftops and fine-grained structural details, suggesting that it better preserves high-frequency geometry while maintaining global shape consistency across multiple viewing angles.
\begin{table}[!ht]
\centering
\caption{Quantitative comparison of SAM 3D and TRELLIS on individual building reconstruction. FID and CMMD are computed for multiple rendered viewpoints per building and averaged accordingly. Lower values indicate better performance.}
\label{tab:quantitative}
\begin{tabular}{cccccc}
\toprule
\multirow{2}{*}{\textbf{Building}} & \multirow{2}{*}{\textbf{View}} & \multicolumn{2}{c}{\textbf{FID} $\downarrow$} & \multicolumn{2}{c}{\textbf{CMMD} $\downarrow$} \\
 &  & \textbf{SAM 3D} & \textbf{TRELLIS} & \textbf{SAM 3D} & \textbf{TRELLIS} \\
\midrule
- & a & 188.30 & 197.60 & 0.79 & 0.79 \\

- & b & 278.65 & 214.86 & 1.61 & 1.61 \\

- & c & 255.56 & 211.90 & 0.79 & 0.79 \\

\textbf{Building1} & \textbf{Avg} & {240.84} & \textbf{208.12} & \textbf{1.06} & \textbf{1.06} \\

- & a & 198.36 & 250.45 & 0.79 & 0.79 \\

- & b & 142.48 & 242.31 & 0.79 & 0.79 \\

- & c & 256.37 & 207.58 & 0.79 & 0.79 \\

\textbf{Building2} & \textbf{Avg} & \textbf{199.07} & 233.45 & \textbf{0.79} & \textbf{0.79} \\

- & a & 266.19 & 264.08 & 0.79 & 0.79 \\

- & b & 255.35 & 174.44 & 0.79 & 0.79 \\

- & c & 218.96 & 324.68 & 2.26 & 0.79 \\

\textbf{Building3} & \textbf{Avg} & \textbf{246.83} & 254.40 & 1.28 & \textbf{0.79} \\

- & a & 214.32 & 283.15 & 1.21 & 1.39 \\

- & b & 263.08 & 262.80 & 0.79 & 1.61 \\

- & c & 183.56 & 170.58 & 1.21 & 1.61 \\

\textbf{Building4} & \textbf{Avg} & \textbf{220.32} & 238.84 & \textbf{1.07} & 1.53 \\

- & a & 181.53 & 349.85 & 0.79 & 2.26 \\

- & b & 179.43 & 243.89 & 1.61 & 3.32 \\

- & c & 218.78 & 259.53 & 0.79 & 1.61 \\

\textbf{Building5} & \textbf{Avg} & \textbf{193.24} & 284.43 & \textbf{1.06} & 2.40 \\

- & a & 286.83 & 320.45 & 0.79 & 0.79 \\

- & b & 211.39 & 178.23 & 0.79 & 0.79 \\

- & c & 250.78 & 247.64 & 0.79 & 0.79 \\

\textbf{Building6} & \textbf{Avg} & 249.66 & \textbf{248.77} & \textbf{0.79} & \textbf{0.79} \\

\textbf{Overall} & \textbf{Avg} & \textbf{224.99} & 244.69 & \textbf{1.01} & 1.22 \\
\bottomrule
\end{tabular}
\end{table}

\subsection{Urban Scene Reconstruction}
\begin{figure*}[!ht]
  \centering
  \includegraphics[width=1.0\textwidth]{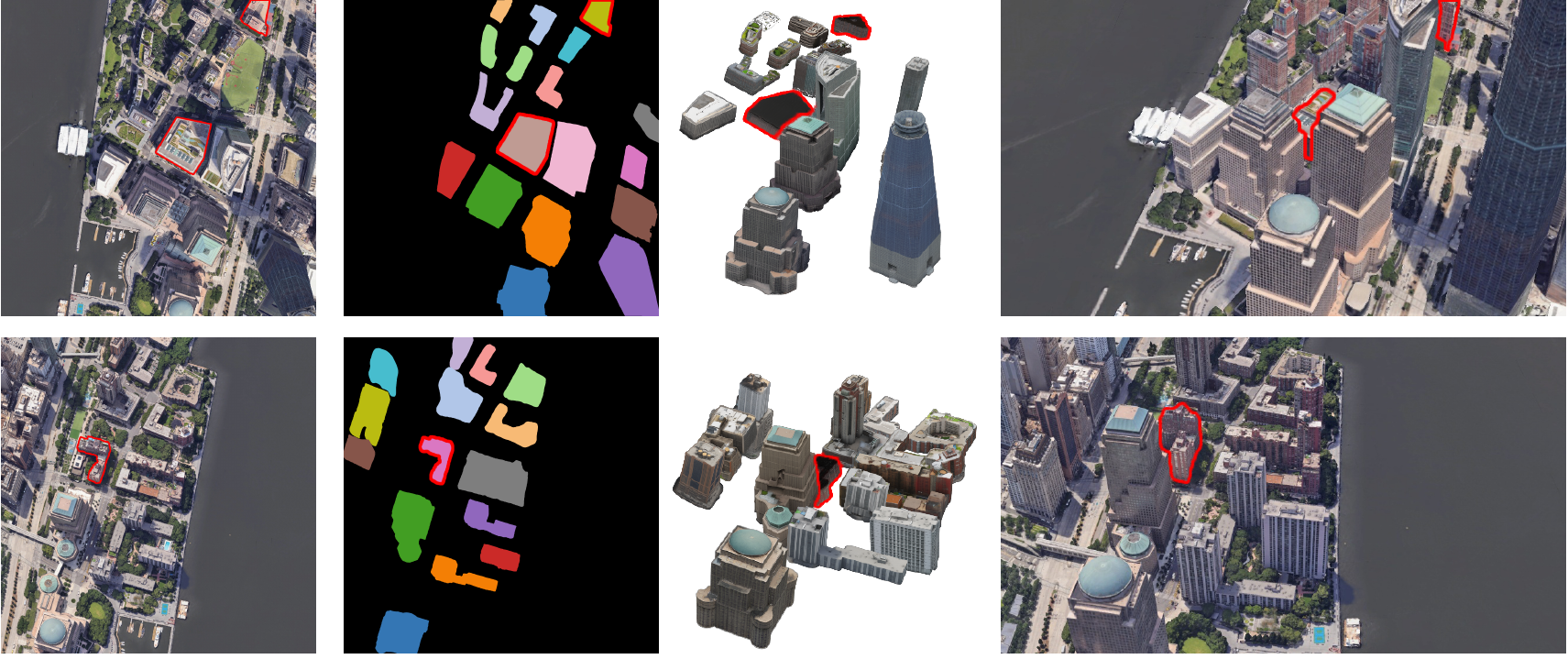}
  \caption{Urban scene reconstruction with SAM 3D. From left to right: remote sensing image, building instance layout, scene reconstruction result, and ground truth. Red boxes indicate buildings with incorrect orientation.}
  \label{fig_3}
\end{figure*}

To assess SAM 3D's applicability to scene-level tasks, we extend our evaluation to urban scene reconstruction using a \emph{segment--reconstruct--compose} pipeline: each building is reconstructed independently and then transformed into a unified coordinate frame via predicted layout parameters. As shown in Fig.~\ref{fig_3}, SAM 3D recovers the overall spatial layout with building footprints and relative positions largely preserved, demonstrating its potential for urban scene reconstruction.

Despite these encouraging results, applying SAM 3D to remote-sensing scene reconstruction reveals several challenges. The object-centric design necessitates independent inference for each building, resulting in computational cost that scales linearly with building count. Moreover, the model's generalization to lower-resolution remote sensing imagery remains to be validated. In terms of scene-level consistency, the lack of inter-object physical constraints may introduce layout drift during composition, and pose-agnostic texture generation can cause orientation ambiguities for rotationally symmetric buildings (see red boxes in Fig.~\ref{fig_3})~\cite{chen2025sam}.

Future work may incorporate scene-level structural priors, adapt to lower-resolution data, and explore multi-object joint prediction to address these challenges.

\section{Conclusion}
This work provides the first systematic evaluation of SAM 3D for monocular remote-sensing 3D building reconstruction. Compared with TRELLIS, SAM 3D achieves superior quality on individual buildings, producing more coherent roof geometry and sharper boundaries. We further extend SAM 3D to urban scene reconstruction via a \emph{segment--reconstruct--compose} pipeline, demonstrating its potential for urban scene modeling.

Despite these promising results, applying SAM 3D to remote-sensing building reconstruction still presents limitations. The object-centric design incurs per-building inference overhead, limiting efficiency for dense urban scenes. Independent object prediction without physical constraints may cause layout drift in composed scenes, and pose-agnostic texture generation can lead to orientation errors for rotationally symmetric buildings.

Future work will focus on integrating scene-level structural priors to improve inter-building consistency, adapting SAM 3D to lower-resolution remote sensing imagery, and exploring multi-object joint prediction to enhance both efficiency and coherence at the scene level.

\small
\newpage
\quad
\newpage
\quad
\newpage
\bibliographystyle{IEEEtranN}

\bibliography{references}

\end{document}